# Feature Selection Based on Orthogonal Constraints and Polygon Area

Zhenxing Zhang[1*], School of Information and Electrical Engineering Ludong University Yantai, China，ludongzx@ldu.edu.cn

Jun Ge[1], School of Compute Science Qufu Normal University, Rizhao, China, gejun_qfnu@163.com

Zheng Wei, School of Compute Science Qufu Normal University, Rizhao, China, Wz0706@qfnu.edu.cn

Chunjie Zhou, School of Information and Electrical Engineering Ludong University Yantai, China，researcherzcj@163.com

Yilei Wang, School of Compute Science Qufu Normal University,Rizhao, China，wang_yilei2019@qfnu.edu.cn

**Abstract**
The goal of feature selection is to choose the optimal subset of features for a recognition task by evaluating the importance of each feature, thereby achieving effective dimensionality reduction. Currently, proposed feature selection methods often overlook the discriminative dependencies between features and labels. To address this problem, this paper introduces a novel orthogonal regression model incorporating the area of a polygon. The model can intuitively capture the discriminative dependencies between features and labels. Additionally, this paper employs a hybrid non-monotone linear search method to efficiently tackle the non-convex optimization challenge posed by orthogonal constraints. Experimental results demonstrate that our approach not only effectively captures discriminative dependency information but also surpasses traditional methods in reducing feature dimensions and enhancing classification performance.

## 1.Introduction

In today's information age, the rapidly increasing scale and complexity of data pose unprecedented challenges to traditional data analysis and machine learning algorithms [1-4]. Feature selection, a crucial research area in data mining, aims to identify the optimal subset of features, reducing the dimensionality of high-dimensional datasets and thereby enhancing the performance of learning algorithms [5-7].

Feature selection methods are commonly categorized into three types: filter, wrapper, and embedded methods [8]. Filter methods evaluate features based on predefined rules or criteria without involving learning algorithms [9]. Examples include information gain (IG) [10], maximum relevance minimum redundancy (mRMR) [11], correlation coefficient (CC) [12], Fisher [13], conditional mutual information maximization criterion (CMIM) [14], and ReliefF [15]. Wrapper methods generate various feature subsets and use learning algorithms to evaluate them, aiming to find the globally optimal subset by maximizing or minimizing an objective function [16].

In recent years, embedded methods have gained widespread attention. Wu et al. [17] introduced a supervised feature selection method, Feature Selection with Orthogonal Regression (FSOR), employing Generalized Power Iteration (GPI) and the Augmented Lagrangian Multiplier method to solve the objective function and evaluate features. Nie et al. [18] developed a Robust Feature Selection (RFS) method that uses the $l_{2,1}$ norm to sparsify the weight matrix, facilitating feature evaluation. In their algorithms, the least squares regression model is employed as the learning algorithm, with the objective function being to find a weight matrix that minimizes the squared residuals between observed values and model predictions [19-23]. To alleviate multicollinearity in the weight matrix and reduce redundancy among features, orthogonal constraints can be introduced into the objective function. However, incorporating these orthogonal constraints converts the optimization of the weight matrix into a non-convex problem, increasing the likelihood of encountering local optima [24,25]. Additionally, the weight matrix represents the correlation between features and labels, serving as parameters that minimize the vertical distance between true and predicted values. Directly using this approach for assessing feature importance might not adequately account for the discriminative dependencies between features and labels.

To overcome these challenges, we propose a novel feature selection method, Polygon Area Feature Selection (PAFS), which leverages polygon areas. First, PAFS employs a hybrid non-monotonic linear search to tackle the non-convex optimization problem posed by orthogonal constraints in the objective function. It optimizes the descent direction to quicken function iteration and carefully selects the step length to stabilize the iteration process. Second, PAFS visualizes the correlation between features and labels as a polygon within the weight matrix, illustrating each feature's contribution to label discrimination. Calculating the polygon's area allows PAFS to integrate the evolving trend and correlation of contributions—namely, the discriminative dependencies between features and labels—into feature importance evaluation, thereby enhancing accuracy.

To verify the effectiveness and accuracy of the PAFS method, preliminary experiments were conducted on eight publicly available datasets and compared with the current advanced feature selection methods. The experimental results show that the PAFS method is effective and superior to the other feature selection methods.

The contributions of this paper include the following:

- Introduction of a novel feature selection method, PAFS, which uniquely uses the area of polygons to represent the weight matrix. This innovative approach ensures a comprehensive evaluation of discriminative dependencies between features and labels, leading to more accurate feature assessment.
- Implementation of a hybrid non-monotonic linear search method to effectively tackle the non-convex optimization problem with orthogonal constraints in the objective function. This method enhances the convergence speed while ensuring stability, thereby optimizing the feature selection process.
- Comprehensive experimental validation on eight publicly available datasets, demonstrating that PAFS outperforms seven other advanced feature selection methods in terms of classification accuracy and dimensionality reduction.

The remainder of this paper is organized as follows: In the 'Related Work' section, this paper delves into the foundational principles of the PAFS method, offering insights into its development. The 'Results' section presents a detailed analysis of our experimental findings, underscoring the efficacy of PAFS compared to existing methods. Finally, the 'Conclusion' section provides a comprehensive summary of our research outcomes, highlighting the practical implications and potential future directions of this study."

## 2. Related work

This section discusses the fundamental principles of the PAFS method. Four main aspects are discussed: the representation of the objective function, the optimization of the objective function, the assessment of feature importance, and the specific workflow of the PAFS method.

First, some symbols $I_n$ denotes an $n \times n$ identity matrix. $1_n = (1,1,1 \cdots 1)^T \in R^{n \times 1}$. $X \in R^{d \times n}$ denotes the data matrix, $Y \in R^{k \times n}$ denotes the label matrix, n represents the number of samples, k represents the number of categories, and d represents the number of features. For any matrix A, the Frobenius norm is defined as $||A||_F^2 = Tr(A^T A)$. For any matrix A, $A^T$ denotes the transpose of the matrix A. For any real number B, $|B|$ denotes the absolute value.

### 2.1 Objective function: Least squares model with orthogonal constraints

General form of the least squares method:

$$\min_{W,b} ||W^T X + b 1_n^T - Y||_F^2 \qquad (1)$$

where $W \in R^{d \times k}$ represents the weight matrix, and $b \in R^{k \times 1}$ denotes the bias vector. The objective of Eq. 1 is to find a suitable W and b that minimize the vertical distance between the actual and predicted values.

The objective function used in this paper integrates orthogonal constraints into the least-squares method, as shown in Eq. 2.

$$\min_{W,b} ||W^T X + b 1_n^T - Y||_F^2 \quad s.t. W^T W = I_k \qquad (2)$$

Compared to the general form, an additional constraint is introduced, namely, the constraint $W^T W = I_k$, which restricts the least-squares model to the Stiefel manifold.

### 2.2 Iterative optimization of the objective function

### 2.2.1 Simplify the objective function

Calculate the partial derivative of Eq. 2 with respect to b, and express b using known symbols.

$$\frac{\vartheta \|W^T X + b 1_n^T\|_F^2}{\vartheta b} = 0$$

It can be seen from the above formula that:

$$b = \frac{1}{n}(Y 1_n - W^T X 1_n)$$

Substituting the above equation into Eq. 2, the objective function is simplified to Eq. 3.

$$\min_W \|W^T X H + Y H\|_F^2 \quad s.t. W^T W = I_k \quad (3)$$

where $H = I_n - \frac{1}{n} 1_n 1_n^T$.

### 2.2.2 Solving the objective function with orthogonal constraints using a hybrid non-monotone linear search method

Restricting the least-squares method to the Stiefel manifold helps to preserve the independence between the features [26]. However, the inclusion of the Stiefel manifold also transforms the optimization of the objective function into a non-convex problem, which makes it difficult to find a global optimal solution.

In this study, a non-monotone linear search method was used to solve the non-convexity introduced by the Stiefel manifold in the least squares model.

Non-monotone linear search method [27-31]: Starting from a point W on the Stiefel manifold (where W is initialized such that $W^T W = I_k$), the direction within the tangent space of W is determined as the search direction F. An suitable step length dt was chosen, whereby iteration is performed within the tangent space, and the iterated points are remapped back onto the Stiefel manifold after completion of the step length. This process was repeated until the objective function converged.

The non-monotone linear search method can accept solutions that do not necessarily decrease the value of the objective function during the search process. This helps to avoid being trapped in the local minima, and this improves the global search capability. In addition, this method allows for dynamic adjustment of the step length, providing greater flexibility. However, this method also has some drawbacks, such as slow convergence speed and oscillations. To address these problems, the non-monotone linear search method was optimized and a hybrid non-monotone linear search method was proposed in this study. This method primarily optimizes the search direction and descent step length. Specifically, a linear combination of three search directions is used in this paper to form a hybrid descent direction. This increases robustness, leads to faster convergence, and overcomes the limitations associated with a single direction. In addition, when projecting points back from the tangent space onto the Stiefel manifold, a hybrid descent step length was formed by averaging the historical and current step lengths. This ensures a smooth convergence and improves the overall convergence performance.

(1) Determine the search direction

To accelerate convergence and overcome directional limitations, a hybrid descent direction consisting of a linear combination of three search directions was chosen in this study:

$$\begin{cases} F_1 = G - W G^T W \\ F_2 = (I_d - W W^T) G \\ F_3 = (\frac{1}{2} I_d - \frac{1}{2} W W^T) G \\ F = \alpha F_1 + \beta F_2 + \gamma F_3 \end{cases} \quad (4)$$

where F1, F2, and F3 represent three directions on the tangent space of W, F denotes the hybrid descent direction, and $\alpha, \beta, \gamma$ are the weights of the search directions, subject to the condition $\alpha + \beta + \gamma = 1$, with $\alpha \geq 0.01, \beta \geq 0.01, \gamma \geq 0.01$. G represents the partial derivative matrix of the objective function with respect to W and $I_d$ denotes a d-dimensional identity matrix.

Proof: The hybrid descent direction F lies within the tangent space of the projection matrix.

Taking the derivative of both sides of the orthogonal constraint, $W^T W = I_k$ with respect to W results in a tangent space at the point W on the Stiefel manifold.

$$T_W St(d, k) = \{Z \in R^{d \times k} : W^T Z + Z^T W = 0\} \quad (5)$$

where $T_W$ represents the tangent space, $St(d, k) = \{W \in R^{d \times k} | W^T W = I_k\}$ denotes the Stiefel manifold, W is the projection matrix, d and k signify the dimensions of W, and Z is a direction within the tangent space of W.

If the search direction lies in the tangent space of W, it should satisfy Eq. 5.

$$W^T F1 + F1^T W = W^T G - W^T W G^T W + G^T W - W^T G W^T W = 0$$

If Eq. 5 is satisfied, F1 lies within the tangent space of the projection matrix.

$$W^T F2 + F2^T W = W^T G - W^T W W^T G + G^T W - G^T W W^T W = 0$$

If Eq. 5 is satisfied, F2 lies within the tangent space of the projection matrix.

$$W^T F3 + F3^T W = \frac{1}{2} W^T G - \frac{1}{2} W^T W W^T G + \frac{1}{2} G^T W - \frac{1}{2} G^T W W^T W = 0$$

If Eq. 5 is satisfied, F3 lies within the tangent space of the projection matrix.

Consequently, F1, F2, and F3 belong to $T_W St(d, k)$. Because $T_W St(d, k)$ is a vector space, it can be concluded that F belongs to $T_W St(d, k)$ by obtaining a linear combination of F1 in which F2 and F3 also belong to $T_W St(d, k)$. This proves that the hybrid descent direction F lies in the tangent space of W and meets the prerequisites for the search direction.

(2) Determine the descent step length

The Barzilai-Borwein method is used to determine the descent step length [32]:

$$a_m^{BB} = \begin{cases} a_m^{BB1} = \frac{\|S_m\|_F^2}{|tr(S_m^T O_m)|} & \text{for all even values of m} \\ a_m^{BB2} = \frac{|tr(S_m^T O_m)|}{\|O_m\|_F^2} & \text{for all odd values of m} \end{cases} \quad (6)$$

where $a_m^{BB}$ represents the Barzilai-Borwein method, $a_m^{BB1}$ denotes the formula for computing the descent step length during even iterations, $a_m^{BB2}$ signifies the formula for computing the descent step length during odd iterations, and m denotes the number of iterations for the objective function starting from m=0. $S_m = W_{m+1} - W_m$ denotes the change in W between two consecutive iterations, while $O_m = \nabla f(W_{m+1}) - \nabla f(W_m)$ represents the gradient change of $f(W_m)$ between the current and the previous iteration. $\nabla f(W) = G - W G^T W$, where $\nabla f(W)$ represents the gradient of the function $f(W)$, and G denotes the partial derivative matrix.

(3) Projecting points back from the tangent space to the Stiefel manifold

Based on Eqs. 1 and 2, the hybrid descent direction F and the step length dt can be determined separately for the points $W_m$ on the Stiefel manifold. Using F and dt, the position of the next point $W_{m+1}$ is determined on the Stiefel manifold.

$$Q_{W_{m+1}} = W \frac{dt_m + dt_{m+1}}{2} F \quad (7)$$

where $Q_{W_{m+1}}$ signifies the position of $W_{m+1}$ in the tangent space, $dt_m$ represents the step length determined by the Barzilai-Borwein at iteration m, $dt_{m+1}$ denotes the step length determined by the Barzilai-Borwein at iteration $m + 1$, and $\frac{dt_m+dt_{m+1}}{2}$ represents the hybrid descent step length (to accelerate convergence and prevent oscillation and divergence). F represents the descent direction at iteration m, and W represents the position of W on the Stiefel manifold at iteration count m.

Projecting the position on the tangent space back onto the Stiefel manifold can be expressed as follows:

$$W_{m+1} = \pi(Q_{W_{m+1}}) = UI_{d,k}V^T \quad (8)$$

where d and k represent the dimensions of the matrix W, U and $V^T$ are derived from the singular value decomposition (SVD) of $Q_{W_{m+1}} \in R^{d \times k}$, where the matrix $Q_{W_{m+1}}$ is decomposed into a d × d unit matrix U, a d × k diagonal matrix $\Sigma$, and a k × k unit matrix $V^T$. $I_{d,k}$ represents the first k rows as $I_k$, and the remaining rows are zero. $Q_{W_{m+1}}$ represents the position of $W_{m+1}$ in the tangent space, $\pi(Q_{W_{m+1}})$ denotes the projection of the position of $W_{m+1}$ in the tangent space back onto the Stiefel manifold, which is the position of $W_{m+1}$.

(4) Algorithm Procedure

---
**Algorithm1** Hybrid Non-monotone Linear Search Algorithm
---
**Input:** Take the initial point W s.t. $W^TW = I_k$
**Initialize parameter:** $dt > 0, 0 < dt_{min} < dt_{max}, \rho, \varepsilon, \mu, \delta \in (0,1), P_0 = 1, m = 0, C_0 = f(W_0)$
**Output:** $W_m$
while $\|\nabla f(W_m)\|_F > \varepsilon$ do

    While $f(\pi(Q_{W_{m+1}})) \geq C_m + \rho\tau Df(W_m)[Y_m(0)]$ do

        $dt = \delta dt$

    end

    Compute: $W_{m+1} = \pi(Q_{W_{m+1}}), P_{m+1} = \mu P_m + 1, C_{m+1} = \frac{\mu P_m C_m + f(W_{m+1})}{P_m} + 1$

    Select step length dt: $dt = a_m^{BB}$

    Update step length dt: $dt = \max(\min(dt, dt_{max}), dt_{min})$, Finally set $m < m + 1$

end

---

## 2.3 Construction of polygonal evaluation features

The dimension of W is d × k, which represents the weight matrix indicating a linear relationship between the features and the target. After convergence of the objective function, the optimal solution for the weight matrix W is obtained. In the following, how to use the discriminative information contained in W is explained for different category recognition tasks, and for comprehensively evaluating the features.

### 2.3.1 Breakdown of the label prediction process

Equation 9 represents the process of predicting the labels, which can be broken down into the solution of the predicted label for the first sample (Eq. 10) and the category equation 1 for the first sample (Eq. 11).

$$W^TX + b = Y$$

$$\begin{pmatrix} W_{11} & \cdots & W_{1d} \\ \vdots & \vdots & \vdots \\ W_{k1} & \cdots & W_{kd} \end{pmatrix} \begin{pmatrix} X_{11} & \cdots & X_{1n} \\ \vdots & \vdots & \vdots \\ X_{d1} & \cdots & X_{dn} \end{pmatrix} + b = \begin{pmatrix} Y_{11} & \cdots & Y_{1n} \\ \vdots & \cdots & \vdots \\ Y_{k1} & \cdots & Y_{kn} \end{pmatrix} \quad (9)$$

$$\begin{pmatrix} Y_{11} \\ Y_{21} \\ \vdots \\ Y_{k1} \end{pmatrix} = \begin{pmatrix} W_{11}X_{11} + b + W_{12}X_{21} + b + \cdots + W_{1d}X_{d1} + b \\ W_{21}X_{11} + b + W_{22}X_{21} + b + \cdots + W_{2d}X_{d1} + b \\ \vdots \\ W_{k1}X_{11} + b + W_{k2}X_{21} + b + \cdots + W_{kd}X_{d1} + b \end{pmatrix} \quad (10)$$

$$Y_{11} = W_{11}X_{11} + b + W_{12}X_{21} + b + \cdots + W_{1d}X_{d1} + b \qquad (11)$$

where, W is the weight matrix, X is the data matrix, Y is the label matrix, and b is the bias. $X_{dn}$ denotes the d-th feature of the n-th sample, $W_{kd}$ represents the weight of the d-th feature in category equation k for all samples, and $Y_{kn}$ represents the probability that the n-th sample belongs to category k.

In this study, the process of solving $Y_{kn}$ was defined as a category equation. For example, Eq. 11 represents category equation 1 for the first sample and is used to determine the probability of the first sample belonging to category 1: A higher value of category equation 1 indicates a higher probability of the first sample belonging to category 1. The column data of the weight matrix W can be defined as the weight that each feature contributes to the category equation, which indicates the contribution of each feature to the different category identification tasks.

A polygonal representation was used to visually illustrate the contribution of the features to the different categories and to enable a comprehensive evaluation of these features. This approach represents the contribution of each feature to different category recognition tasks, with the aim of achieving an effective feature assessment.

### 2.3.2 Representation of the polygon

To facilitate the use of integration to calculate the areas of the polygons, the converged weight matrix W was first subjected to quadrant processing, limiting the elements in W to the first quadrant. $W_\Delta$, $W_\Delta \in R^{d \times k}$ can be used to obtain a new weight matrix. If each row of data in $W_\Delta$ is drawn as a polygon, it results in a total of d polygons, where each polygon corresponds to a feature that represents the weight exerted by each feature across the k category equations. The polygon diagrams for different numbers of classifications are shown in Figs. 1–4.

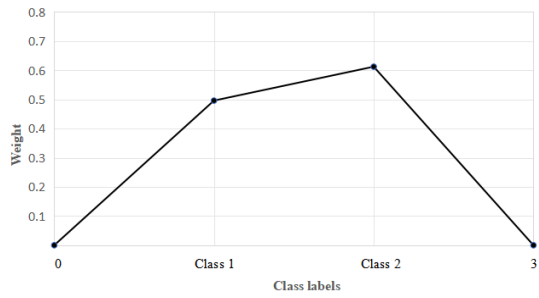

**Fig.1** Polygon plot for binary classification

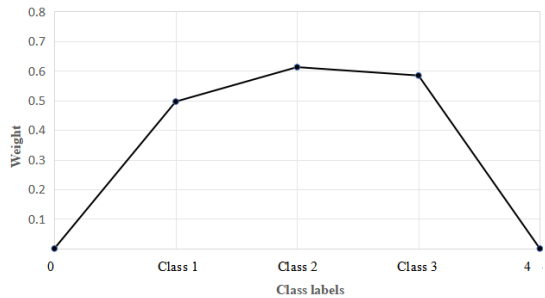

**Fig.2** Polygon plot for ternary classification

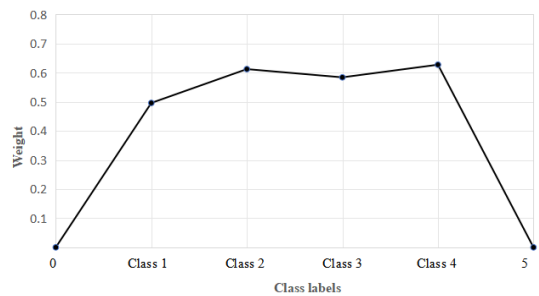

**Fig.3** Polygon plot for quaternary classification

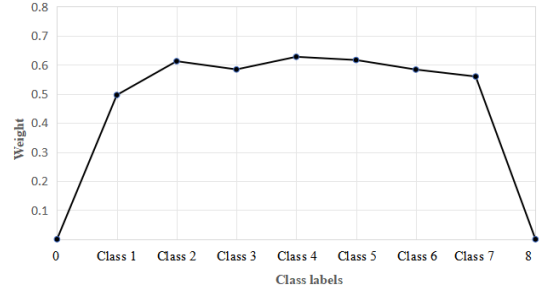

**Fig.4** Polygon plot for septenary classification

In Figs. 1–4, the x-axis represents the class labels with two points, 0 and k + 1, added to simplify the calculation. The y-axis represents the weights of the features in the class equations.

### 2.3.3 Calculating the area of the polygon

Step 1: Find a straight line between every two adjacent points.

Step 2: The evaluation of feature d is based on the shape enclosed by a straight line and the x-axis. The formula for the evaluation is as follows:

$$P_j = \int_0^1 a_0 x + b_0 dx + \int_1^2 a_1 x + b_1 dx + \cdots + \int_k^{k+1} a_k x + b_k dx \qquad (12)$$

where j = 1, 2, ..., d (where d represents the number of features), k is the number of categories, a is the slope of the line, and b is the line bias. $P_j$ denotes the evaluation value of the j-th feature.

Each feature was assigned an evaluation value P, where a higher P value indicates a higher evaluation of the feature. The sorting of the P-values results in an overall evaluation of the features.

### 2.4 Process of feature selection with the PAFS method

(1) Different values of αβγ are taken and the hybrid non-monotonic linear search algorithm is run until convergence. The descent curve (as shown in Fig. 5) and the area value S enclosed by the x- and y-axes during each iteration are calculated. A smaller area indicates that fewer iterations are required for convergence, and a steeper descent occurs. The values of αβγ that correspond to the minimum area are determined as follows:

$$\min_{\alpha,\beta,\gamma} S_i = \int_0^1 a_0 x + b_0 dx + \int_1^2 a_1 x + b_1 dx + \cdots + \int_{c-1}^c a_{c-1} x + b_{c-1} dx \qquad (13)$$

where i = 0, 1, 2, ..., s-1, and s represents all possible values for α, β, γ, c denotes the number of iterations required to converge using the non-monotonic linear search algorithm, a signifies the slope of the line, and b represents the bias.

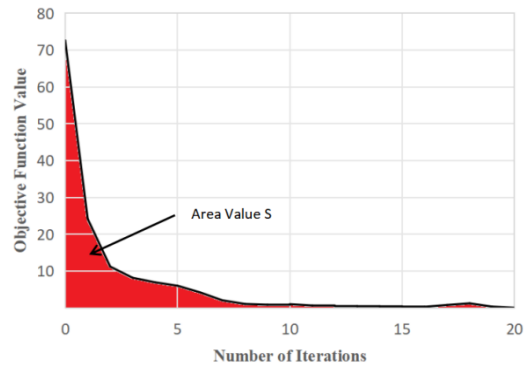

**Fig.5** Descent curve of iterations

(2) Determine the projection matrix W corresponding to the optimal αβγ values. For each feature, polygons are generated and their respective areas are computed and ranked to derive the feature evaluation results.

**Algorithm2** PAFS Method

**Input:** the data matrix $X \in R^{d*n}$, the label matrix $Y \in R^{k*n}$

**Initialize:** $W \in R^{d*k}$ satisfying $W^T W = I_k$

**Output:** Overall evaluation of features

**Repeat s times:**

    Update W using Algorithm 1 until convergence, preserving the matrix W after each round of convergence.

**Retrieve the parameters $\alpha\beta\gamma$ that satisfy $\min_{\alpha,\beta,\gamma} S_i$, along with the optimal projection matrix W corresponding to these parameters:**

$$min_{\alpha,\beta,\gamma} S_i = \int_0^1 a_0 x + b_0 dx + \int_1^2 a_1 x + b_1 dx + \cdots + \int_{c-1}^c a_{c-1} x + b_{c-1} dx$$

**Using the optimal W, draw polygons for each feature and calculate the area $P_j$ □□□ □□□□:**

$$P_j = \int_0^1 a_0 x + b_0 dx + \int_1^2 a_1 x + b_1 dx + \cdots + \int_k^{k+1} a_k x + b_k dx$$

**Sort Pj** (sorted by magnitude, where larger P values indicate higher weight, indicating the feature's advantageousness for classification, and vice versa).

## 3. Results

To validate the accuracy and effectiveness of the newly proposed PAFS method, eight publicly available datasets were selected for experimental verification. Detailed information about the datasets is listed in Table 1.

Table 1    Public Dataset Details

| Dadaset | Samples | Features | Classes |
|---|---|---|---|
| Vehicle | 846 | 18 | 4 |
| Segment | 2310 | 19 | 7 |
| Chess | 3196 | 36 | 2 |
| Control | 600 | 60 | 6 |
| Uspst | 2007 | 256 | 10 |
| Binalpha | 1404 | 320 | 36 |
| Corel_5k | 5000 | 423 | 50 |
| Yeast | 1484 | 1470 | 10 |

For eight publicly available datasets, the PAFS method was first used to evaluate the features within each dataset. The features were then arranged in descending order based on their scores. Taking the Segment dataset as an example, Figure 6 and Table 2 illustrate the process of evaluating features using the PAFS method.

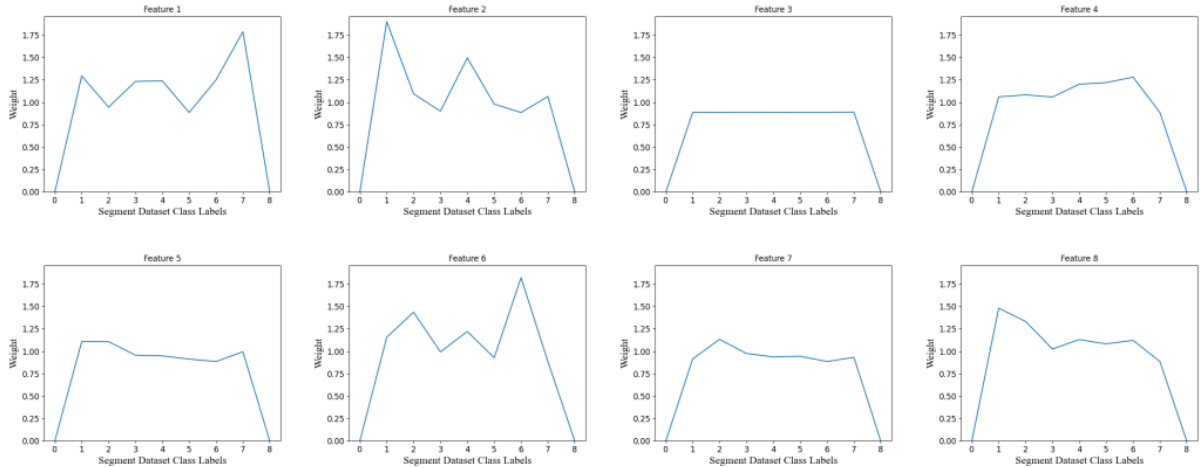

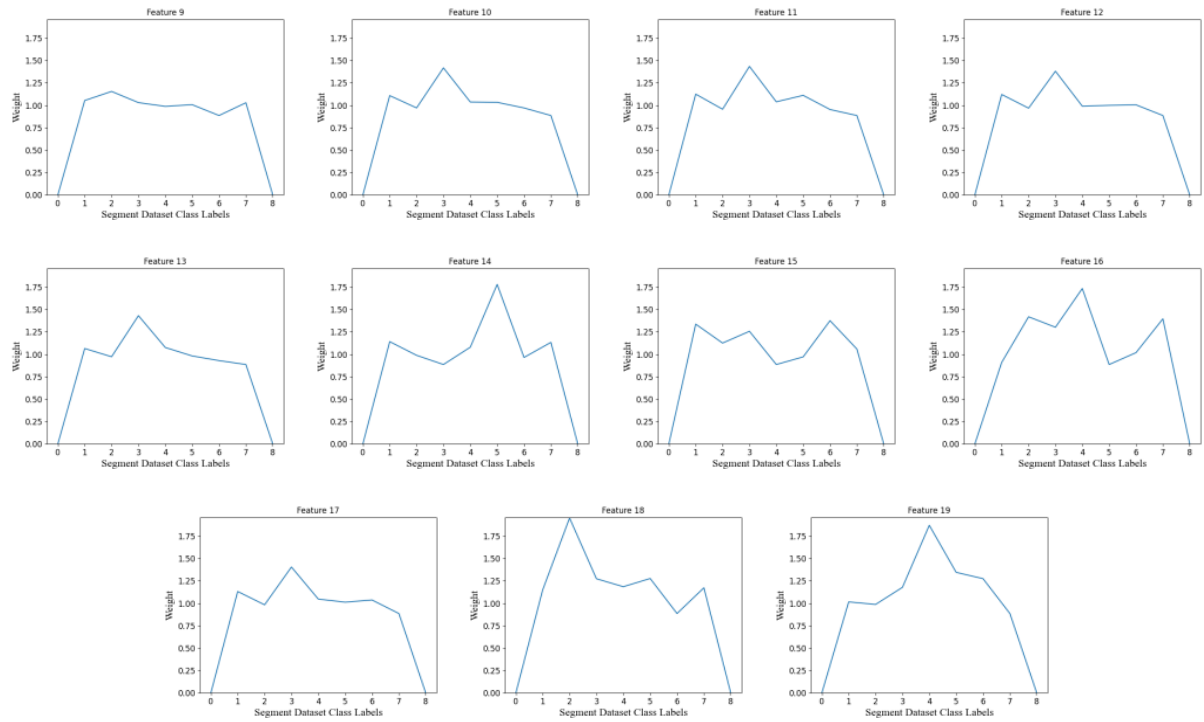

**Fig.6** Polygonal image plotted for the segment dataset

**Table 2** Area sorting table of polygonal images for the segment dataset

| Segment Dataset | |
| --- | --- |
| Feature Index | Polygon Area |
| 17 | 8.90140718011748 |
| 15 | 8.66932216394962 |
| 0 | 8.640768072790518 |
| 18 | 8.562538739035004 |
| 5 | 8.446889671869293 |
| 1 | 8.327717794631546 |
| 7 | 8.066805674647107 |
| 14 | 8.014162200371079 |
| 13 | 7.976757591125824 |
| 3 | 7.794208854350149 |
| 10 | 7.507410732672485 |
| 16 | 7.506183980614221 |
| 9 | 7.427295685567 |
| 11 | 7.354029499133743 |
| 12 | 7.345599308109161 |
| 8 | 7.151561136639092 |
| 4 | 6.919125015340341 |
| 6 | 6.728524013369444 |
| 2 | 6.213175593641908 |

  A backward feature selection method was then applied, in which features with lower scores were progressively eliminated to form a new subset of features, which were then used to train a support vector machine (SVM) classifier (linear). The subset of features that yielded the highest recognition rate was considered the optimal feature subset.

  Training Parameters: Seventy percent of the data from the feature subset was used for training, whereas the remaining 30% was used for testing. The training data was subjected to 100 training iterations within an SVM (linear) classifier. The average recognition rate on the test set across these iterations was taken as the experimental result.

The above experiment was repeated and the results tested for all feature subsets. The experimental results for the eight datasets are shown in Fig. 7.

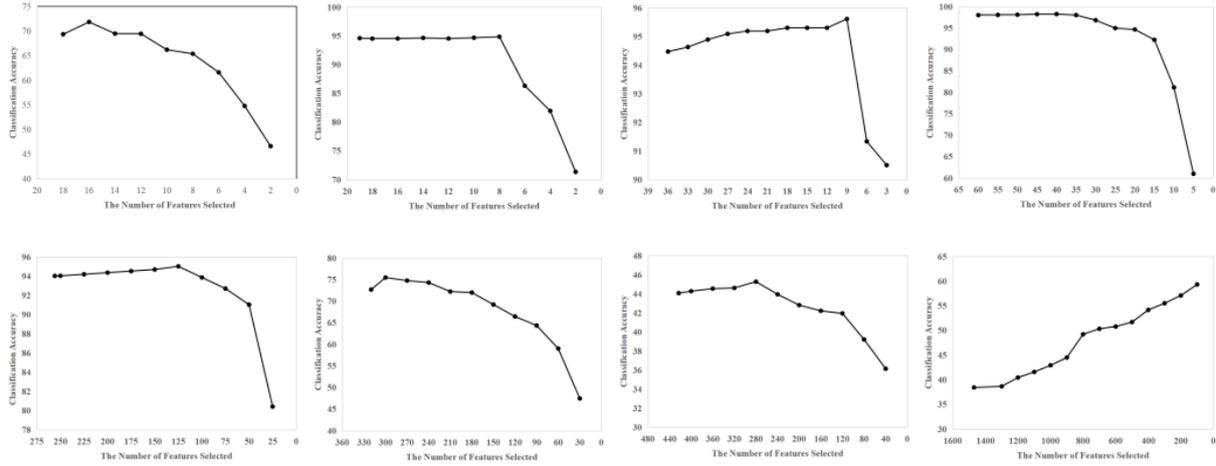

**Fig.7** Test results on 8 datasets

From Fig. 7, it can be visually observed that the recognition rates of the PAFS + SVM method first increase and then decrease as the number of features decreases. In the PAFS method, the features are ranked, with the features at the bottom indicating lower scoring and redundant features. Removing these features the performance of the model, reduces the risk of overfitting, and accelerates the training speed, resulting in an upward trend in recognition rates. Once the features are reduced to a certain number, the remaining features are informative, task-relevant, and exhibit low intercorrelations. Further deletion of features shows a decreasing trend in the recognition rate.

Effectiveness: In the Vehicle dataset, the highest recognition rate was achieved when there were 16 features present, indicating that this set of 16 features formed the optimal feature subset for recognizing the Vehicle dataset. In the Segment dataset, the highest recognition rate occurred when there were eight features present. In the Chess dataset, the highest recognition rate occurred when there were nine features. In the Control dataset, the highest recognition rate was achieved when there were 40 features present. For the Uspst dataset, the highest recognition rate occurred when there were 125 features present. In the Binalpha dataset, the highest recognition rate occurred when there were 300 features present. In the Corel_5k dataset, the highest recognition rate was achieved when there were 280 features present. In the Yeast dataset, the highest recognition rate was achieved when there were 100 features present. The number of features in the optimal feature subset was reduced compared to the original feature set, demonstrating the effectiveness of the PAFS method.

Accuracy: By using the PAFS method to rank features, the features at the forefront are those with the highest information content and are the most relevant to the prediction task. Preserving these features helps to keep the recognition rate. When the number of features in the Chess dataset was reduced from 36 to 3, the recognition rate only dropped by 4%, and was still over 90%. In the Corel_5k dataset, the recognition rate only dropped by 8% when the number of features was reduced from 423 to 40. This proved the accuracy of the evaluation results of the PAFS method.

Furthermore, experiments on objective function iterations were conducted to demonstrate that the hybrid non-monotone linear search method can achieve faster and more stable convergence. The results are listed in Table 3.

**Table 3** Comparison of the convergence performance of the hybrid non-monotone linear search method and the non-monotone linear search method is conducted through the area value S

| Dataset | Non-monotone Linear Search Method | Hybrid Non-monotone Linear Search Method |
| --- | --- | --- |
| Vehicle | 24.61 | 14.52 |
| Segment | 446.24 | 371.17 |
| Chess | 104.02 | 35.50 |
| Control | 16.85 | 9.12 |
| Uspst | 39.79 | 23.09 |
| Binalpha | 10.49 | 5.55 |
| Corel_5k | 184.19 | 67.84 |
| Yeast | 1.65 | 1.00 |

The results show that the hybrid non-monotone linear search method has a smaller area under the loss function curve for all eight datasets. This indicates fewer iterations of the objective function and a steeper descent process, indicating the superiority of the hybrid non-monotone linear search method over the non-monotone linear search method.

In addition comparative experiments on the recognition rates of the optimal feature subsets were conducted in this study. The PAFS method was compared with seven other advanced feature selection methods: ReliefF, Fisher, CMIM, mRMR, CC, IG, and FSOR. An experiment was conducted using eight publicly available datasets. In each experiment, the dataset was divided into a training set and a testing set in a 7:3 ratio and they were placed in SVM (linear), SVM (RBF), K-nearest neighbors (KNN), and random forest (RF) classifiers to test the classification results. Determining the recognition rate of the optimal feature subset for each method was the objective. Each experiment was repeated 100 times, and the average value was used as the final classification recognition rate. The comparative results are listed in Table 4.

**Table 4** Compares the average classification accuracy of 8 feature selection methods across eight publicly available datasets

| Dataset | Vehicle | | | |
|---|---|---|---|---|
| Accuracy | SVM(Linear) | SVM(RBF) | KNN | RF |
| ReliefF | 69.16 | 72.90 | 68.11 | 72.04 |
| Fisher | 66.07 | 66.00 | 64.17 | 68.50 |
| CMIM | 66.21 | 66.39 | 63.63 | 69.59 |
| mRMR | 69.16 | 67.98 | 61.81 | 71.25 |
| CC | 67.98 | 66.40 | 62.20 | 70.47 |
| IG | 66.79 | 66.00 | 61.41 | 68.50 |
| FSOR | 69.37 | 73.08 | 68.18 | 72.29 |
| PAFS | **71.82** | **73.49** | **69.27** | **73.47** |
| Dataset | Segment | | | |
| Accuracy | SVM(Linear) | SVM(RBF) | KNN | RF |
| ReliefF | 89.87 | 94.64 | 93.90 | 94.38 |
| Fisher | 87.59 | 92.20 | 93.36 | 93.07 |
| CMIM | 92.06 | 95.23 | 94.51 | 96.10 |
| mRMR | 91.77 | 95.38 | 94.80 | 95.81 |
| CC | 87.87 | 91.05 | 91.48 | 92.20 |
| IG | 84.41 | 93.65 | 91.91 | 93.01 |
| FSOR | 92.16 | 95.54 | 95.34 | 96.17 |
| PAFS | **94.84** | **96.79** | **96.02** | **97.99** |
| Dataset | Chess | | | |
| Accuracy | SVM(Linear) | SVM(RBF) | KNN | RF |
| ReliefF | 93.63 | 95.76 | 94.25 | 95.65 |
| Fisher | 93.21 | 95.09 | 94.05 | 95.31 |
| CMIM | 93.63 | 95.82 | 94.78 | 95.72 |
| mRMR | 93.52 | 95.72 | 94.36 | 96.05 |
| CC | 93.63 | 95.40 | 94.05 | 95.09 |
| IG | 93.63 | 96.03 | 94.46 | 95.51 |
| FSOR | 93.84 | 96.13 | 94.92 | 96.25 |
| PAFS | **95.61** | **97.70** | **94.27** | **96.57** |
| Dataset | Control | | | |
| Accuracy | SVM(Linear) | SVM(RBF) | KNN | RF |
| ReliefF | 91.11 | 92.77 | 95.55 | 91.53 |
| Fisher | 90.11 | 92.22 | 95.00 | 92.22 |
| CMIM | 84.44 | 92.12 | 93.33 | 91.66 |
| mRMR | 91.00 | 91.66 | 94.44 | 91.11 |
| CC | 85.55 | 92.22 | 92.22 | 90.00 |
| IG | 81.11 | 92.22 | 91.66 | 90.00 |
| FSOR | 91.48 | 92.88 | 94.69 | 91.36 |
| PAFS | **98.25** | **94.44** | **97.77** | **98.88** |
| Dataset | Uspst | | | |
| Accuracy | SVM(Linear) | SVM(RBF) | KNN | RF |
| ReliefF | 91.69 | 94.35 | 92.02 | 91.52 |
| Fisher | 91.36 | 93.02 | 91.69 | 90.03 |
| CMIM | 92.54 | 95.11 | 92.50 | 92.90 |
| mRMR | 91.19 | 94.68 | 93.02 | 92.19 |
| CC | 90.53 | 93.85 | 91.19 | 89.36 |
| IG | 91.36 | 93.18 | 91.02 | 90.53 |
| FSOR | 92.83 | 95.54 | 93.13 | 93.01 |
| PAFS | **95.01** | **95.73** | **94.31** | **93.55** |
| Dataset | Binalpha | | | |
| Accuracy | SVM(Linear) | SVM(RBF) | KNN | RF |
| ReliefF | 65.74 | 66.89 | 62.03 | 64.58 |
| Fisher | 61.80 | 63.65 | 60.80 | 59.02 |
| CMIM | 68.53 | 69.53 | 65.24 | 67.66 |
| mRMR | 63.88 | 66.20 | 62.50 | 62.96 |
| CC | 63.65 | 64.81 | 59.49 | 62.50 |
| IG | 62.96 | 64.12 | 61.80 | 61.11 |
| FSOR | 68.25 | 72.07 | 65.46 | 67.83 |
| PAFS | **75.46** | **73.60** | **68.17** | **71.00** |
| Dataset | Corel_5k | | | |
| Accuracy | SVM(Linear) | SVM(RBF) | KNN | RF |
| ReliefF | 40.26 | 41.53 | 29.86 | 41.00 |
| Fisher | 36.93 | 38.50 | 34.60 | 32.53 |
| CMIM | 40.31 | 42.76 | 33.45 | 42.13 |
| mRMR | 40.61 | 43.16 | 31.44 | 42.15 |
| CC | 38.93 | 39.53 | 27.46 | 38.13 |
| IG | 36.50 | 38.90 | 27.53 | 38.06 |
| FSOR | 40.60 | 43.03 | 32.53 | 42.33 |
| PAFS | **45.26** | **44.66** | **33.60** | **43.86** |
| Dataset | Yeast | | | |
| Accuracy | SVM(Linear) | SVM(RBF) | KNN | RF |
| ReliefF | 48.70 | 42.55 | 38.90 | 40.11 |
| Fisher | 32.13 | 42.47 | 31.46 | 34.60 |
| CMIM | 55.11 | 61.66 | 54.53 | 58.90 |
| mRMR | 42.11 | 57.45 | 43.10 | 60.11 |
| CC | 42.47 | 60.89 | 45.16 | 57.30 |
| IG | 57.07 | 58.65 | 56.17 | 58.42 |
| FSOR | 58.04 | **63.35** | **56.98** | 60.46 |
| PAFS | **59.32** | 63.14 | 56.62 | **64.49** |

The results show that the PAFS method has a significant numerical advantage in terms of the average classification accuracy over the other seven feature selection methods in the eight publicly available datasets,. In most cases, the feature subset selected by the PAFS method achieved the highest recognition rate. This indicates that the feature selection method proposed in this study selects the most effective feature subset and provides accurate feature evaluation.

## 4. Conclusion

In this study a novel feature selection method based on polygon area was proposed. This method projects feature vectors onto a smaller subspace, effectively preserving the discriminative dependencies in the data. It uses a polygonal image to comprehensively assess the contribution of features in different category recognition tasks, thereby providing a more integrated feature evaluation. A non-monotone linear search method on the Stiefel manifold is used and optimized to theoretically demonstrate the feasibility of the PAFS method. The effectiveness and accuracy of this method was experimentally validated using eight publicly available datasets. In conclusion, this study has shown that the PAFS method is a viable approach whose accuracy outperforms other feature selection algorithms.